\documentclass[english, letterpaper, 10pt]{ieeeconf}

\makeatletter
\IEEEtriggercmd{\reset@font\normalfont\scriptsize}
\makeatother
\IEEEtriggeratref{1}

\newcommand{\subparagraph}{}

\usepackage[english]{babel}
\usepackage[utf8]{inputenc}

\usepackage{color,xcolor,ucs}

\usepackage{floatrow}
\usepackage{tabularx}
\usepackage{float}
\usepackage{amsfonts}
\usepackage{helvet}         
\usepackage{courier}        
\usepackage{type1cm}        
\usepackage{amsmath}
\usepackage{dsfont}
\usepackage{amssymb}
\usepackage{makeidx}         
\usepackage{comment}         
\usepackage{graphicx}        
\usepackage{multicol}        
\usepackage[bottom]{footmisc}
\usepackage{bm}

\usepackage{cite}
\usepackage{url}

\usepackage{xr-hyper}

\usepackage{epsfig}
\usepackage{epstopdf}
\usepackage{caption}
\usepackage{subcaption}

\usepackage{xspace}
\usepackage{rotating}

\usepackage{graphicx}
\usepackage{threeparttable}
\usepackage{multirow}
\usepackage[font=scriptsize,labelfont=bf]{caption}

\usepackage[margin=0.75in,bottom=0.85in]{geometry}

\usepackage{algpseudocode,algorithm,algorithmicx}
\DeclareMathOperator*{\argmax}{arg\,max}

\definecolor{yakov}{rgb}{0.0,0.0,1.0}
\definecolor{vova}{rgb}{0.0,0.,0.0}

\newcommand{\VT}[1]{{\color{vova} #1}}

\algrenewcommand\algorithmicrequire{\textbf{Input:}}
\algnewcommand{\LineComment}[1]{\State \(\triangleright\) #1}

\newcommand{\appropto}{\mathrel{\vcenter{
			\offinterlineskip\halign{\hfil$##$\cr
				\propto\cr\noalign{\kern2pt}\sim\cr\noalign{\kern-2pt}}}}}

\title{DUQIM-Net: Probabilistic Object Hierarchy Representation for Multi-View Manipulation}

\author{Vladimir Tchuiev, Yakov Miron, and Dotan Di Castro 
\thanks{The authors are with the Bosch Center for Artificial Intelligence (BCAI), Haifa, Israel. Authors' email addresses: {\tt \{Vladimir.Tchuiev, Yakov.Miron, Dotan.DiCastro\}@il.bosch.com}.
}
}

\date{}

\graphicspath{{figures/}}

\makeatletter
\def\endthebibliography{%
	\def\@noitemerr{\@latex@warning{Empty `thebibliography' environment}}%
	\endlist
}
\let\NAT@parse\undefined
\makeatother

\usepackage[colorlinks,citecolor=red,urlcolor=blue,bookmarks=false,hypertexnames=true]{hyperref}

\begin{document}
	
	\maketitle
	
	\thispagestyle{empty}
	\pagestyle{empty}

\begin{abstract}
    Object manipulation in cluttered scenes is a difficult and important problem in robotics. To efficiently manipulate objects, it is crucial to understand their surroundings, especially in cases where multiple objects are stacked one on top of the other, preventing effective grasping. We here present DUQIM-Net, a decision-making approach for object manipulation in a setting of stacked objects. In DUQIM-Net, the hierarchical stacking relationship is assessed using Adj-Net, a model that leverages existing Transformer Encoder-Decoder object detectors by adding an adjacency head. The output of this head probabilistically infers the underlying hierarchical structure of the objects in the scene. We utilize the properties of the adjacency matrix in DUQIM-Net to perform decision making and assist with object-grasping tasks. Our experimental results show that Adj-Net surpasses the state-of-the-art in object-relationship inference on the Visual Manipulation Relationship Dataset (VMRD), and that DUQIM-Net outperforms comparable approaches in bin clearing tasks. 
\end{abstract}
	
	
	
	\section{Introduction}
	\label{sec:introduction}
	

Robotic manipulation in scenes with clustered objects is an important challenge in robotics, with the most fundamental task being grasping \VT{(e.g. {\cite{Morrison19icra}})}. Robotic object grasping is a cornerstone in many robotic tasks such as bin-clearing and target-oriented grasping-the-invisible, which have wide applications, e.g., in industrial assembly, search and rescue and domestic appliances. To effectively perform manipulation tasks, the robot must accurately perceive its surroundings, which is a significant challenge in environments with object clutter. To accomplish its task in such environments, the robot must possess a semantic understanding of the stacking relationships between the objects present. Inferring these relationships, however, is a difficult problem due to heavy occlusions. Therefore, a robust robot perception and manipulation approach is required.

\begin{figure}[h!]
    \begin{center}
        \includegraphics [width=0.8\textwidth]{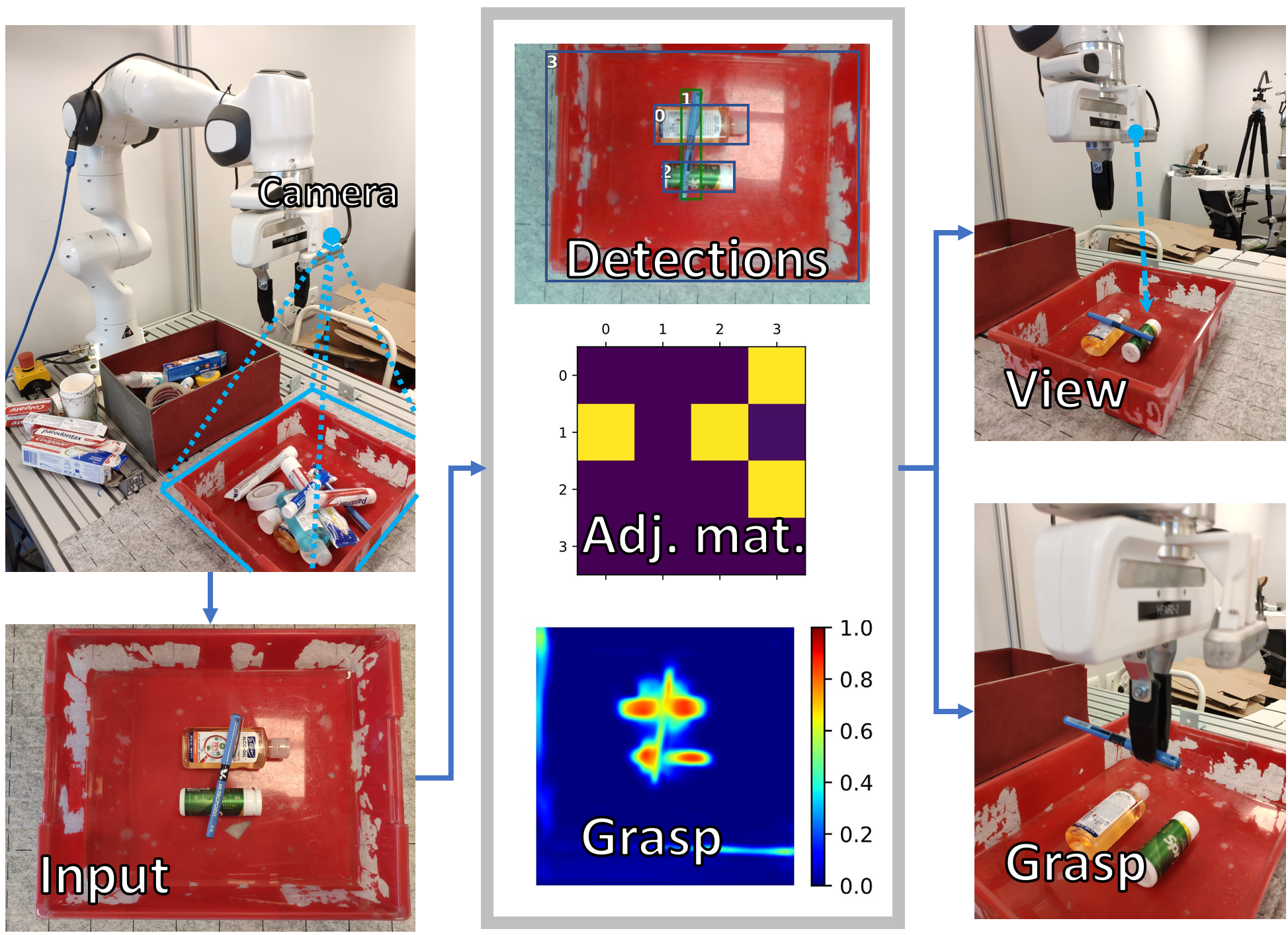}
    \end{center}
    \caption{The problem setting is an environment with cluttered objects. In our approach, DUQIM-Net, the hierarchical relationships between the detected objects is inferred via an adjacency matrix (Adj. mat. in the figure) from an RGB image taken by an arm-mounted camera. In addition, the system infers a grasping quality map. Using the inferred data, a decision whether to grasp an object or to view the clutter from another viewpoint is made.}
    \label{fig:Diagram_fig_high}
\end{figure}

Understanding the stacking relationships between objects, i.e., the stacking hierarchy, is crucial for determining the manipulation order for a certain environment. Between two given objects, we consider whether one is placed directly on top of the other. All the object relationships in the scene can be aggregated into a directed graph, which can be utilized to infer the manipulation order. However, inferring this graph from a single image introduces uncertainty to the hierarchy inference, and may require a multi-view approach to reduce this uncertainty (\cite{Morrison19icra, Lin20sj}). On the one hand, current state-of-the-art deep-learning-based approaches \VT{(e.g. \cite{Zuo21ijcnn})} assume the hierarchy based on a single image and do not take into account the uncertainty in object hierarchy. These approaches are, therefore, error prone and do not inherently support multi-view inference. On the other hand, more classical model-based approaches do not generalize well for novel scenarios and scenes \VT{(see \cite{Rosman11ijrr})} We propose a data-driven deep-learning-based approach that probabilistically infers the object hierarchy in a scene and enables data-fusion from multiple views of the scene.

In this work, we represent the hierarchy graph as a weighted graph \emph{adjacency matrix}. This adjacency matrix representation opens the possibility of fusing stacking information from multiple images, thus allowing safer manipulation, especially in scenarios with \emph{fragile} objects whose dropping might be dangerous. We further propose an action-selection scheme that takes advantage of this representation. Other potential applications may include exploiting the adjacency matrix's sparsity to deduce how difficult it will be to clear a given cluttered scene or identify the object that carries the largest number of objects on top of it, for maximum impact when performing a push action on the clutter.


\VT{
The paper's main contributions are as follows: We propose DUQIM-Net (\textbf{D}ue-order \textbf{U}nderstanding for \textbf{Q}uality \textbf{I}nference and \textbf{M}anipulation Network), a system for object manipulation within a cluttered environment based on RGB images alone. At the core of our approach is Adj-Net (Adjacency-Network), a method for probabilistic object hierarchy detection as a weighted directed graph adjacency matrix based on Transformer Encoder-Decoder architecture. Using Adj-Net, we outperform the state-of-the-art on the VMRD dataset \cite{Zhang18ichr} in terms of inferring the object hierarchy. In DUQIM-Net we exploit the probabilistic nature of the adjacency matrix to both facilitate an incremental update for reasoning about the environment, and to perform efficient action selection in situations where the grasping order is crucial.
}


    
    
    
    

In the paper, we first present the mathematical background for our adjacency matrix and describe some useful properties that can be exploited for robotic manipulation. Then, we present DUQIM-Net, with Adj-Net and the action-selection segment. Then, we compare the performance of Adj-Net to the state-of-the-art approaches for stacking hierarchy inference on the Visual Manipulation Relationship Dataset (VMRD \cite{Zhang18ichr}). Finally, we present the advantage of utilizing Adj-Net in a bin-clearing task on a real Franka Panda Emika robot. This paper is accompanied by supplementary material \cite{Tchuiev22ral_sup}, which provides further details and results.

\section{Related Work}

\textbf{Object Detection:} The most basic ability of our system is to accurately detect objects in the environment and draw a bounding box around them. This has been an active research topic in recent years. In the last decade, with the rise of deep-learning algorithms, all state-of-the-art object detection approaches are deep-learning-based. Notable algorithms include R-CNN \cite{Girshick14cvpr} and the followups Fast R-CNN \cite{Girshick15iccv} Faster-R-CNN \cite{Ren16ieee}, all YOLO versions \cite{Redmon16cvpr} and their many variations. These algorithms require bounding box priors, which are created based on segmentation for R-CNN and its variants, or on the division of the image into separate segments in the case of YOLO. Algorithms like R-CNN consider a multi-step operation to produce accurate bounding boxes, while algorithms like YOLO are single-step approaches that sacrifice inference accuracy for computational speed. The recent development of Transformer Encoder-Decoder algorithms \cite{Vaswani17nips} has led to the inception of DETR \cite{Carion20eccv} and later Deformable DETR \cite{Zhu20arxiv}, which do not use bounding box priors and produce impressive results. In this paper, we utilize Deformable DETR as the backbone for our Adj-Net segment.

\textbf{Object Hierarchy Inference:} Early methods for object hierarchy inference were model-based and made some assumptions on the objects' shapes. For example, Rosman et al. \cite{Rosman11ijrr} utilizes a model that infers object contact points from a point cloud to interpret the object hierarchy. Conversely, Sjoo et al. \cite{Sjoo11iros} learns spatial relationships of simulated objects using a Sparse Bayesian approach. For the specific domain of various industrial parts, Oh et al. \cite{Oh12case} proposed a pipeline for picking objects in a clutter with a stereo input using geometric pattern matching. Mojtahedzadeh et al. \cite{Mojtahedzadeh13iros} leveraged classical mechanics, combined with spatial inference, to infer the stacking relationship of boxes in a clutter. Panda et al. \cite{Panda16jirs} utilized neural networks for object detection and segmentation; the spatial relationship is inferred in a heuristic manner via the objects' center of mass. These model-based approaches, however, fail to generalize well for challenging conditions and/or novel objects and require depth measurements. In contrast, our approach is data-driven and is capable of generalizing to many configuration scenarios, objects and scenes. Moreover, it requires only RGB image data as input.

In recent years, several data-driven approaches based on deep-learning have been proposed. With the introduction of the VMRD dataset \cite{Zhang18ichr}, which contains real-life object stacking scenes with annotated relationships, several methods were proposed for stacking hierarchy inference. Multi-task CNN by Zhang et al. \cite{Zhang19iros}, followed by VMRN by Zhang et al. \cite{Zhang20prl}, are CNN-based approaches that were trained on VMRD. Zuo et al. \cite{Zuo21ijcnn} proposed GVMRN, a hybrid network of CNN and GNN (Graph Neural Network). Future papers may use the REGRAD simulated image dataset \cite{Zhang21arxiv} to train models that infer object hierarchy effectively. In contrast to these methods, our proposed approach utilizes a Transformer Encoder-Decoder architecture to perform the task. In addition, our approach reasons about the uncertainty in object hierarchy inference.

A related problem to object hierarchy inference is unseen object instance segmentation, where a model attempts to perform instance segmentation on occluded parts of an object in an image in addition to the visible parts. This problem, with accompanying datasets, was introduced by Zhu et al. \cite{Zhu17cvpr}. UOIS 2D \cite{Xie20pmlr} and 3D \cite{Xie21tro} and UOAIS-Net \cite{Back21arxiv} utilize RGB-D images to do so. As with the current works on stacking hierarchy inference, these works do not reason about the uncertainty concerning the stacking hierarchy.

	
	\section{Adjacency Matrix Preliminaries}
	\label{sec:preliminaries}
	
\subsection{Weighted Directed Graph Hierarchy Representation}

Consider an environment with multiple stacked objects. These object are stacked in a hierarchical structure, with some supported by the the objects below them. This structure can be represented as a graph $\mathcal{G} \triangleq \{ \mathcal{V}, \mathcal{E}\}$ with $N_{\mathcal{V}}$ vertices $\nu \in \mathcal{V}$ and $N_{\mathcal{E}}$ edges $\epsilon \in \mathcal{V}$. The vertices represent the objects themselves, while the directed edges represent the \emph{direct} hierarchical relationship between these objects. In the case, e.g., of objects $a$ and $b$, if object $a$ is placed \emph{directly} above object $b$, then $\mathcal{G}$ contains vertices $\nu_a$ and $\nu_b$ that are connected via directed edge $\epsilon_{a \rightarrow b}$. 

In this paper, the object hierarchy is unknown and must be inferred from observations of incomplete data. We propose to address this by representing the inferred hierarchy as a weighted directed graph $\mathcal{G} \triangleq \{ \mathcal{V}, \mathcal{E}, \mathcal{W} \}$, where each edge $\epsilon \in \mathcal{V}$ has an associated weight $\omega \in \mathcal{W}$, with values between 0 and 1. This weight represents the probability that the relationship described by the edge exists, i.e.:
\begin{equation}\label{eq:Weight_Def}
    \omega \triangleq \mathbb{P}(\exists \epsilon | I),
\end{equation}
where $I$ is the image measurement from which the hierarchy graph is inferred. In an ideal setting, the value of $\omega$ is either 0 or 1 for all $\omega$.

\subsection{Graph Adjacency Matrix}

The weighted directed graph can be represented as an adjacency matrix $A$. In the case of $N_\mathcal{V}$ vertices, the adjacency matrix is a square $N_\mathcal{V} \times N_\mathcal{V}$, where its diagonal values are 0 and the off-diagonal elements are the weights $\omega$ of all the edges in $\mathcal{E}$ that connect between all $\nu \in \mathcal{V}$, i.e.:
\begin{equation}\label{eq:Adj_Def}
    A \triangleq
    \left[\begin{array}{cccc}
        0 & \omega_{12} & \cdots & \omega_{1 \mathcal{V}} \\
        \omega_{21} & 0 & \cdots & \omega_{1 \mathcal{V}} \\
        \vdots & \vdots & \ddots & \vdots \\
        \omega_{\mathcal{V}1} & \omega_{\mathcal{V}2} & \cdots & 0
    \end{array}\right] \in \mathbb{R}^{N_{\nu} \times N_{\nu}}.
\end{equation}
This weighted graph adjacency matrix representation has useful properties that can be utilized in object manipulation, detailed in the following subsection. An example of a weighted adjacency matrix is presented in Fig.~\ref{fig:Adj_eg}.

\begin{figure}
     \centering
      \begin{subfigure}[b]{0.40\textwidth}
         \centering
         \includegraphics[width=\textwidth]{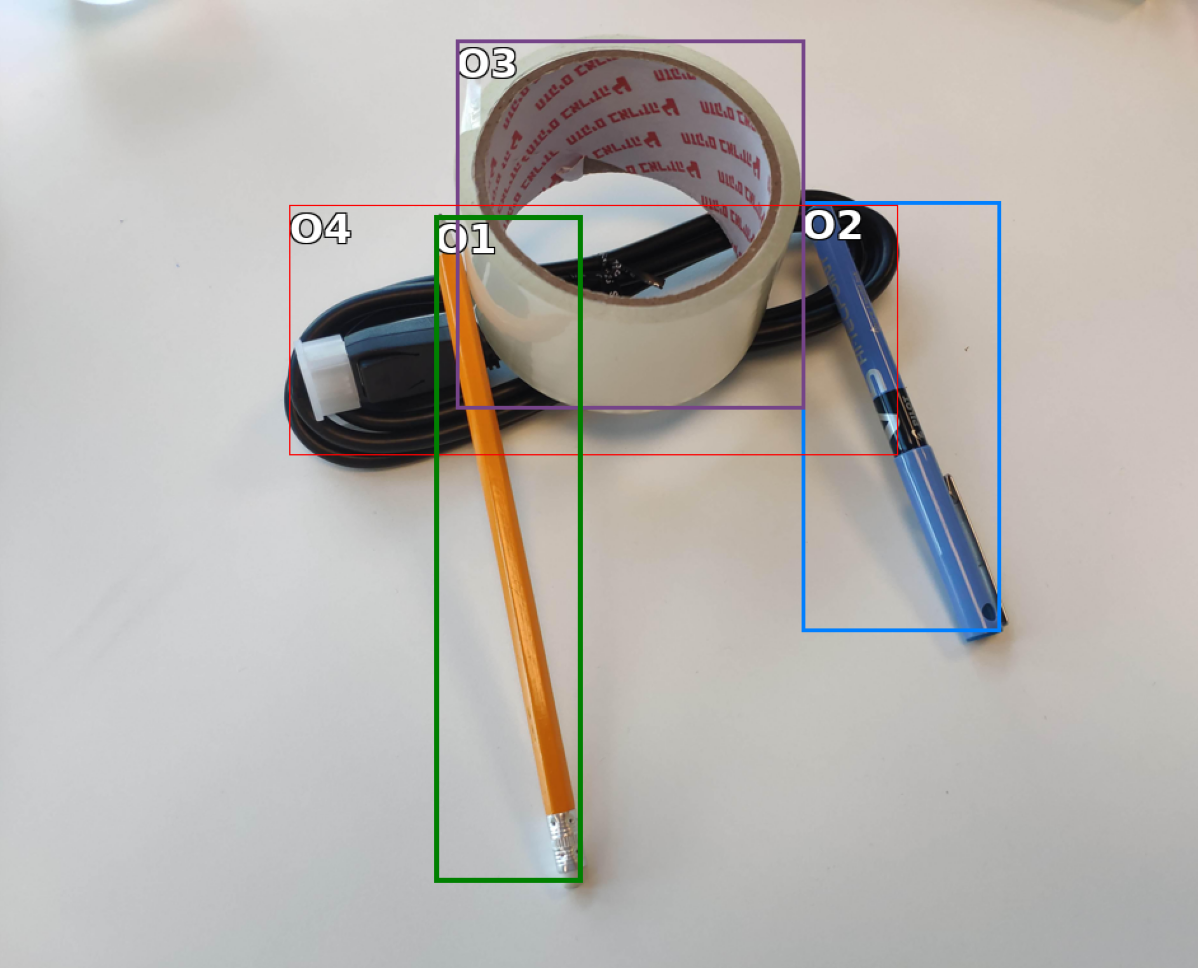}
         \caption{Example Image}
         \label{fig:three sin x}
     \end{subfigure}
     \begin{subfigure}[b]{0.36\textwidth}
         \centering
         \includegraphics[width=\textwidth]{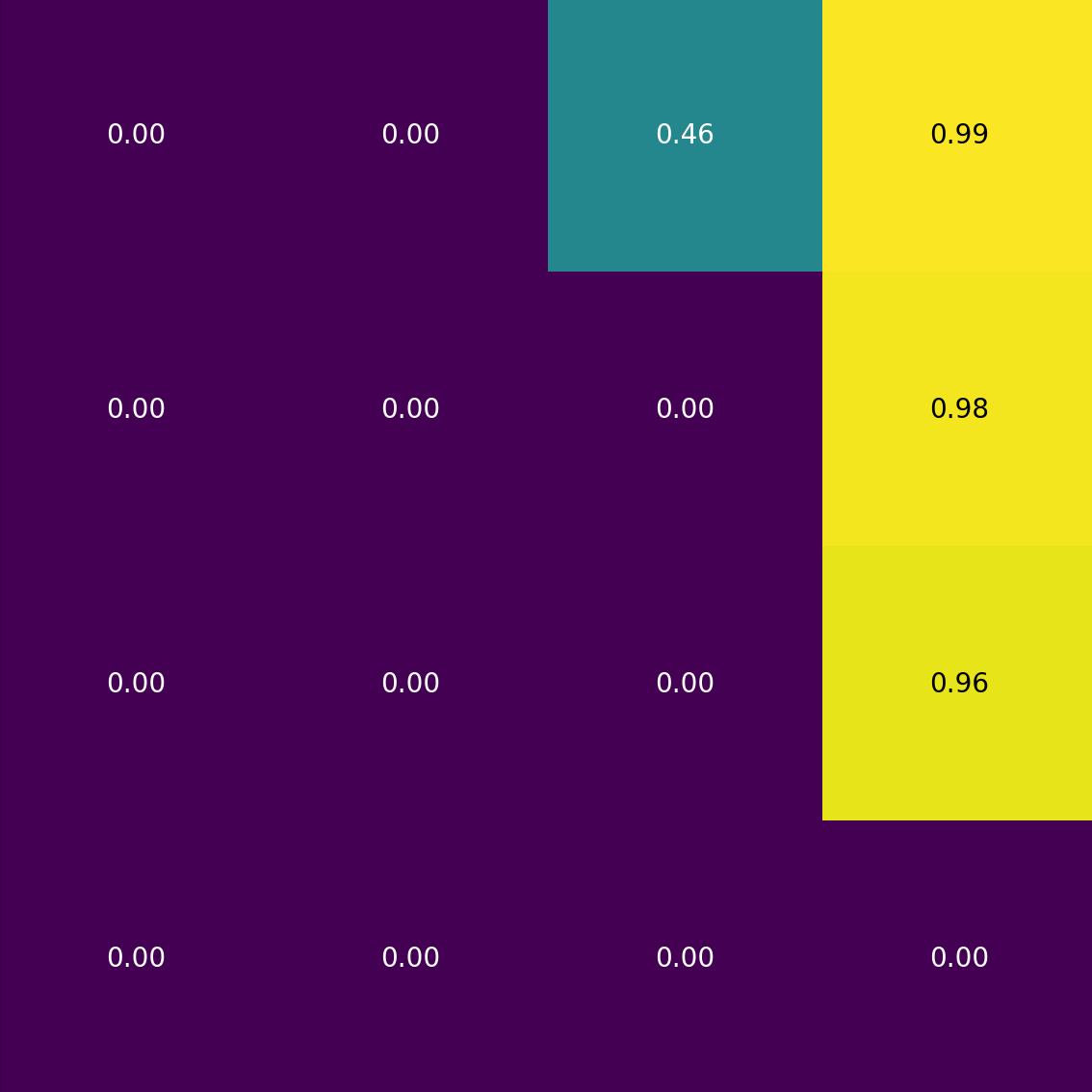}
         \caption{Adjacency Matrix}
         \label{fig:y equals x}
     \end{subfigure}
    \caption{Example of adjacency matrix inference. In \textbf{(a)}, an image is passed through an object detector, which produces and indexes bounding boxes. In \textbf{(b)}, the corresponding inferred adjacency matrix $A$ is presented, with probabilities of object relationships, where the probability of object $i$ being placed over object $j$ is the element $A_{ij}$. The rows and columns correspond to the object numbering in \textbf{(a)}.}
\end{figure}\label{fig:Adj_eg}

\subsection{Adjacency Matrix Properties}

\subsubsection{Adjacency Matrix Moments}
The $n$th moment of the adjacency matrix, i.e., $A^n$, represents the probability that the objects are connected by an $n$th degree connection. For example, consider three objects, $a$, $b$, and $c$. Object $a$ is placed on $b$, and $b$ is placed on $c$. Then the element that corresponds to $a$ over $c$ in $A^2$, i.e., $(A_{ac})^2$, which represents the probability that object $a$ is stacked on $c$ via one intermediary object between them, in this case, object $b$. This property is useful, for example, when performing a push that affects the majority of objects within a scene, as it allows the detection of the object that supports the greatest number of objects, directly and indirectly.

\subsubsection{Safe Grasping Probability}
Conversely, to infer the safest grasp, we can determine the probability that a given object does not support any other object. For object $i$, we denote this probability as $\mathbb{P}(\neg \epsilon_i|I)$. For all the objects, the probability can be computed in a batch (\cite[Sec.~1]{Tchuiev22ral_sup}):
\begin{equation}\label{eq:Safest_Grasp_Eq}
    \log(\mathds{1}_{N_{\mathcal{V}} \times N_{\mathcal{V}}} - A^T) \cdot \mathds{1}_{N_{\mathcal{V}}} =
    \log \mathbb{P}(\neg \epsilon|I)
\end{equation}
where $\mathbb{P}(\neg \epsilon|I) \triangleq 
\left[ \mathbb{P}(\neg \epsilon_1|I) ,\cdots, \mathbb{P}(\neg \epsilon_{N_{\mathcal{V}}}|I) \right]^T$, and $\mathds{1}$ denotes a vector or a matrix of ones, where the subscript represents its dimensions. We can use this vector of probabilities, e.g., to determine the safest object to grasp with the highest $\mathbb{P}(\neg \epsilon_i|I)$, or all the objects that are "safe" to grasp that surpass a threshold on $\mathbb{P}(\neg \epsilon_i|I)$.

\subsubsection{Mutli-View Data Fusion}
A single RGB image might not suffice to correctly infer the underlying hierarchical relationships of an object in a scene, so a multi-view data fusion approach should be considered. The probabilistic nature of $A$ naturally enables inference from multiple measurements. Using Bayesian inference, we can compute a posterior adjacency matrix $A^{post}$ from a set of images $H$, each $I$ with its own $A(I)$ matrix. Assuming that the object ordering in all the $A$ matrices is identical, each element $(i,j)$ of the posterior is computed via the Bayes Rule as follows (\cite[Sec.~2]{Tchuiev22ral_sup}):
\begin{equation}\label{eq:A_Fusion_Raw}
    A^{post}_{ij} \triangleq \mathbb{P}(\exists \epsilon_{ij}|H) = 
    \frac{\prod_{I \in H}A_{ij}(I)}
    {\prod_{I \in H} (1 - A_{ij}(I)) +  \prod_{I \in H} A_{ij}(I)}.
\end{equation}
%
Keeping the object ordering between all $A(I)$ requires to accurately solve the data association problem between the objects in different images, which is outside the scope of this paper. In our experiments, we utilize an existing approach to solve this problem, detailed in Sec.~\ref{sec:experiments}.

	\section{DUQIM-Net}
	\label{sec:approach}
	Given an RGB image, obtained from an arm-mounted camera on a robotic arm, of a cluttered scene with multiple \VT{rigid or soft}, possibly fragile or sensitive, objects. We aim to infer the next action the robot should perform to \emph{safely} achieve its manipulation objective, e.g., bin-clearing or grasping-the-invisible. Fig.~\ref{fig:Diagram_fig_high} represents the high-level architecture of our system. At its core, the system utilizes a Transformer Encoder-Decoder-based object detector (e.g. \cite{Carion20eccv, Zhu20arxiv}) to get bounding box proposals and to probabilistically infer the spatial relationship between the objects they represent. In addition, the system infers from the RGB image a grasping affordance map. Using the detection results, the adjacency matrix and the affordance map, the system infers the action the robot should take, allowing for an action selection approach that reasons about the objects' hierarchical relationship. The useful properties of the adjacency matrix open the possibility of reasoning about the quality of the inference, as well as of fusing multi-view data from multiple RGB images.

\begin{figure*}
    \begin{center}
        \includegraphics [width=0.75\textwidth]{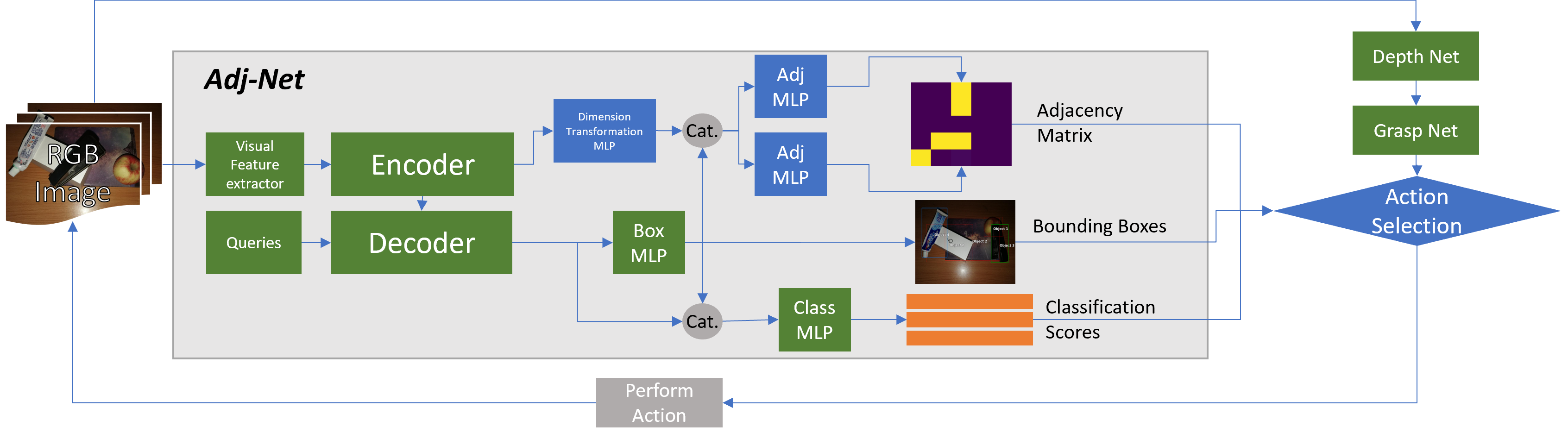}
    \end{center}
    \caption{High-level diagram of DUQIM-Net. The blue blocks denote our core contributions. The input RGB images pass through Adj-Net, which produces bounding box coordinates. The coordinates are concatenated with the decoder output and are fed into the classification MLP, which produces the classification scores as output. The encoder output passes through an MLP for dimension reduction, is concatenated with the bound box coordinates, and then passes through the adjacency head MLP to form the adjacency matrix. In addition, a grasping affordance map is produced in conjunction with depth estimation. All the information is fed into an action-selection segment, whose output is an action command. Finally, the action is executed and a new RGB image is taken. The presented image is part of the VMRD dataset \cite{Zhang18ichr}.}
    \label{fig:Diagram_fig_high}
\end{figure*}

\subsection{Adj-Net}

Adj-Net (\textbf{Adj}acency \textbf{Net}work) forms the core of the system. The detector and the adjacency head it utilizes infer the objects' bounding box, class, and adjacency matrix, to be used by the action selection. We propose an architecture expansion that uses off-the-shelf object detectors to spot the objects and infer the spatial relationship between them.

\subsubsection{Architecture}

Adj-Net's architecture consists of four parts: encoder-decoder backbone, bounding box head, classification head and adjacency head. 

The detector's encoder outputs $e$ feature vectors with hidden size $h$. Per standard Transformer Encoder-Decoder architecture, these feature vectors serve as keys and values that are input into the decoder. The detector's output is $q$ feature vectors of size $h$, corresponding to $q$ decoder queries. We denote the encoder's feature vector output as $x_e \in \mathbb{R}^{h \times e}$, and the decoder's as $x_d 
\in \mathbb{R}^{h \times q}$.

By passing $x_d$ through a Multi-Layer-Perceptron (MLP), each feature vector in $x_d$ is translated into a bounding-box proposal $x_b \in \mathbb{R}^4$. Subsequently, the bounding box coordinates are concatenated with $x_d$ to form $x_d' \doteq x_d \bigcup x_b \in \mathbb{R}^{(h+4) \times q}$; $x'_d$ is passed through the classification head to get the corresponding $q$ vectors of candidate class probabilities. The classification head is an MLP that receives $x'_d$ as an input, and for $c_t$ candidate classes yields $(c_t+1)$ outputs, with the last one corresponding to the "empty" class.

The decoder's output correlates to the bounding box proposals from the queries, carrying information about both the bounding box coordinates and classification. Therefore, the decoder output dilutes the visual feature correlations from the encoder. To counter this problem, and thus improve the accuracy of the adjacency matrix inference, we directly utilize the information in $x_e$ to create the adjacency matrix. Our aim here is that the output will be a $q \times q$ matrix, which corresponds to all the pairs of the bounding box proposals. First, $x_e^T$ is passed through an MLP block that outputs $\chi_e \in \mathbb{R}^{h \times q}$. This MLP block serves a dual purpose: First, its output dimension coincides with the dimensions of $x_b$ such that the bounding box location information is taken into account. Second, this MLP takes into account the connections between encoder feature relationships, attempting to connect them to the hierarchy between the objects. As with $x'_d$, $\chi_e$ is then concatenated with $x_b$ to form $\chi'_e$. Because the adjacency matrix describes the relationships between stacked objects, its values must be agnostic to the objects' ordering. Namely, for different orderings of objects, the corresponding elements in the adjacency matrices must be identical. Therefore, the network must include a matrix multiplication operation to ensure the output is ordering agnostic. Our proposed adjacency head consists of two MLP segments that preserve the input dimension of $\chi'_e$, whic is passed through the adjancecy head's MLP segments, and then a matrix multiplication is performed to infer a preliminary adjacency matrix $A_p \in \mathbb{R}^{q \times q}$. In the following equation, we formulate the operation. Consider $g(\cdot)$ to be the dimension reduction MLP block, and $f_1(\cdot)$ \& $f_2(\cdot)$ to be the adjacency head segments. Then, the total operation performed on $x_e$ is:
%
\begin{eqnarray}
    \chi'_e &=& g(x^T_e) \cup x_b \\
    A_p &=& \sigma \left( \frac{f_1(\chi'_e)^T 
    \cdot f_2(\chi'_e)}{h} \right).
\end{eqnarray}
The element-wise sigmoid $\sigma(\cdot)$ operation is necessary to keep $A_p$'s values between 0 and 1. Dividing the multiplied matrix by $h$ assists in keeping the values close to 0, thus reducing the vanishing gradient problem of the sigmoid activation function \cite{Hochreiter98ijufk}.

Inspired by \cite{Carion20eccv}, the final results are determined by the classification probability vector. Each bounding box whose corresponding class is "unknown" is discarded, as is, subsequently, its row and column in the preliminary adjacency matrix. In the end, the classification vectors determine $N_\mathcal{V}$ bounding boxes out of $q$ proposals, and the size of the final adjacency matrix is $N_{\mathcal{V}} \times N_{\mathcal{V}}$.

\subsubsection{Training}

Adj-Net is trained in a supervised manner from a training set that consists of RGB image inputs and labels that consist of the ground-truth classes, bounding boxes, and spatial relationships. Two available online examples for such datasets are VMRD \cite{Zhang18ichr} and REGRAD \cite{Zhang21arxiv}.
Given a labeled image, Adj-Net produces raw bounding box proposals, classification scores and an adjacency matrix. To compute a loss, the inferred bounding box proposals must be matched to the "closest" target bounding box and class. Inspired by DETR \cite{Carion20eccv}, we use the Hungarian Algorithm \cite{Kuhn55nrlq} to match bounding boxes during training and, furthermore, to infer the relevant rows and columns in $A_p$ and form the matched adjacency matrix $A_m$.

The classification head is trained via a weighted cross-entropy loss. For image $I$ with proposals $p \in \mathcal{P}$ and class $c$, the weighted cross-entropy loss is:
\begin{equation}\label{eq:CE_Loss}
 \mathcal{L}_c = - \sum_{p \in \mathcal{P}} \sum_{c=1}^{c_t+1} w_c \cdot \mathbb{P}_{gt}(c_p|I) \log{\mathbb{P}(c_p|I)}.
\end{equation}
As the number of detections for which the ground-truth class is "unknown" is generally significantly larger than the number of objects in the scene, the "unknown" class is weighted by a small $w_{c_t+1} \ll 1$, while for the other classes, $w_c=1$.

For the matched bounding boxes, we use a linear combination of $l_1$ loss $\mathcal{L}_{l_1}$ and a Generalized Intersection over Unions (GIoU) loss $\mathcal{L}_{GIoU}$ \cite{Rezatofighi19cvpr}, as used by Carion et al. \cite{Carion20eccv}. 

The adjacency matrix $A_m$ (and, by extension, $A_p$) is generally a sparse matrix with many 0 values and few 1 values, Adj-Net risks converging during learning into the convenient local minimum of a matrix of zeros. This is a significant problem when using element-wise $l_1$ and $l_2$ losses, as they do not significantly penalize a false 0. This problem is exacerbated when trying to learn the entire $A_p$ padded with zeros or ones for the rows and columns that are not matched. For this reason, we use binary element-wise cross-entropy loss over $A_p$:
\begin{equation}\label{eq:Adj_Loss}
    \mathcal{L}_{adj} = - \sum_{i,j \in A_m} \left( A_{ij}^{gt} \log{A_{ij}} + (1- A_{ij}^{gt}) \log{(1- A_{ij})} \right),
\end{equation}
where $A_{ij}$ is the $(i,j)$th element of $A_m$, and $A_{ij}^{gt}$ is the ground truth of that element. Training over the target $A_m$ is slower, as only a subset of the $q \times q$ matrix is trained. However, it avoids the predicted adjacency matrix converging to a prediction of a matrix of zeros, as the relative sparsity of $A_m$ (i.e., the ratio of elements with value 0 to all the elements in the matrix) is much lower than that of $A_p$. To offset the slower training, and to generalize well to unseen objects in general, we utilize the super-convergence learning rate scheduling technique \cite{Smith19isop} in conjunction with the AdamW optimizer \cite{Loshchilov17arxiv}.

The overall loss function is a weighted sum of all the individual loss functions:
\begin{equation}\label{eq:Tot_Loss}
    \mathcal{L}_{total} = \gamma_{c} \mathcal{L}_c + \gamma_{l_1} \mathcal{L}_{l_1} + 
    \gamma_{GIoU} \mathcal{L}_{GIoU} + \gamma_{adj} \mathcal{L}_{adj},
\end{equation}
where all the $\gamma$s are hyperparameters.

\subsection{Action Selection}\label{sec:Action_Select}

We propose an action selection algorithm that takes advantage of Adj-Net. The action selection decides whether to grasp or take another viewpoint. Grasping is performed by moving the robotic arm to the designated position and an appropriate angle for grasping the object, opening the grasping claws, and grabbing the object. Taking another viewpoint is done via a command to the robot to observe a designated point.

From Adj-Net, we get as inputs the reduced adjacency matrix $A_m$ and the corresponding bounding box coordinates. In addition, the action selection block receives as input a grasp quality map with a consideration whether to grasp or not. Many recent approaches for the production of such maps rely on depth information, which can be measured via RGB-D cameras or estimated via various depth estimation techniques \cite{Godard17cvpr, Fu18cvpr}. As our system processes RGB images, we resort to the latter approach, specifically, monocular depth estimation techniques based on deep learning (e.g., DPT-Net \cite{Ranftl21iccv}).

To take advantage of $A$, we need to determine how safe it is to grasp an object. To do so, we compute $\mathbb{P}(\neg \epsilon|I)$ given $A$ using Eq.~\eqref{eq:Safest_Grasp_Eq}, and then we compute the maximal information entropy of $\mathbb{P}(\neg \epsilon|I)$, denoted $\mathcal{H}_{max}(A)$:
\begin{equation}\label{eq:Entropy}
\begin{split}
    \mathcal{H}_{max}(A) &\triangleq \max_i ( -
    \mathbb{P}(\neg \epsilon_i|I) \cdot \log(\mathbb{P}(\neg \epsilon_i|I))\\-
    &(1 - \mathbb{P}(\neg \epsilon_i|I)) \cdot \log(1 - \mathbb{P}(\neg \epsilon_i|I))
    ).
\end{split}
\end{equation}
$\mathcal{H}_{max}(A)$ plays an integral part in our action-selection switching algorithm and is responsible for determining whether the robot should take an additional viewpoint. For a world state tuple of robot states and environment $\mathcal{E}$, we consider the inferred bounding box coordinates $\mathcal{C}_{bb}$ ,entropy threshold $\mathcal{H}_{th}$, grasp quality map $\mathcal{Q}$ with a threshold for quality of the value $q_{th}$, and information-gain map for taking another viewpoint $\mathcal{M}$. 

The algorithm for the action selection is presented in Alg.~\ref{alg:ASA}. Given a world state $\mathcal{E}$ and the hyperparameter $\mathcal{H}_{th}$, we take an image, then pass it through Adj-Net and the depth and grasping networks to obtain the adjacency matrix and the corresponding grasping map. In addition, we compute a map of the information gained if taking additional views of the scene. If $\mathcal{H}_{max}(A) \leq \mathcal{H}_{th}$, we perform a grasp, else $\mathcal{H}_{max}(A) \geq \mathcal{H}_{th}$, we take another viewpoint. In both cases, we perform the best action in terms of quality or information gain.

\begin{figure}[]
    \begin{center}
        \includegraphics [width=0.9\textwidth]{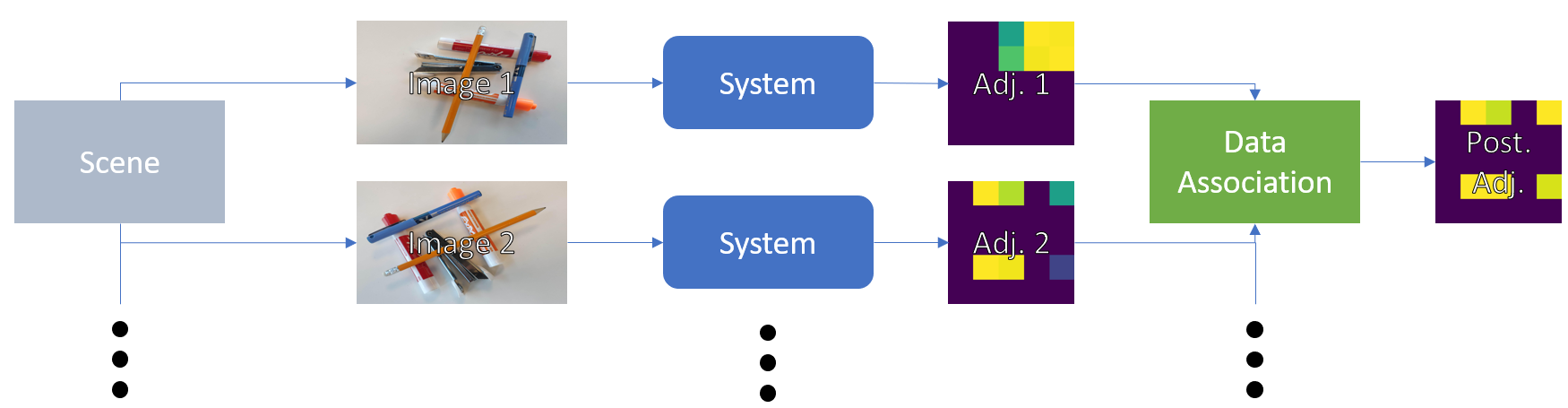}
    \end{center}
    \caption{The multi-view pipeline for posterior adjacency matrix computation. From a scene, we take two or more images, then pass them through the system and get sets of bounding box predictions and adjacency matrices. We match the bounding boxes with the data association algorithm and get a posterior adjacency matrix.}
    \label{fig:Multi-View_Pipeline}
\end{figure}

If the algorithm selects to go to another viewpoint, we run Adj-Net again on the new image from the updated $\mathcal{E}$, match the predicted bounding boxes of the previous image with those of the current image, and infer a posterior adjacency matrix using Eq.~\eqref{eq:A_Fusion_Raw}. As we aim to select viewpoints predicted to provide a high information gain, the posterior adjacency matrix will describe the observed scene with low uncertainty. This pipeline is depicted in Fig.~\ref{fig:Multi-View_Pipeline}. The size of the latest posterior adjacency matrix is determined by the adjacency matrix from the latest image; only the entries that correspond to matching bounding boxes from the previous images are updated. Using the posterior adjacency, action selection is performed again. The action selection is repeated until the task is complete or it reaches a failure state (e.g., 10 consecutive grasping failures).

\begin{algorithm}
\footnotesize
\caption{Action selection algorithm}\label{alg:ASA}
\begin{algorithmic}[1]
    \Require $\mathcal{E}$, $\mathcal{H}_{th}$ \Comment{Gets a view and switching parameters}
    \State{$a \gets \emptyset$}
    \State $A, \mathcal{C}_{bb} \gets \text{AdjNet}(\mathcal{E})$ \Comment{Run Adj-Net}
    \While{$a \notin \textbf{grasp}$} \Comment{View until grasp}
        \State{$\mathcal{Q}, \mathcal{M} \gets \text{Nets}(\mathcal{E})$}
        \Comment{Run DUQIM-Net}
        \State $\mathcal{Q}_{bb} \gets \mathcal{Q}(\mathcal{C}_{bb})$ 
        \Comment{Mask grasp quality map}
        \If{$\mathcal{H}_{max}(A) \leq \mathcal{H}_{th}$}
            \State $a \gets \{ \textbf{grasp: } \argmax(\mathcal{Q}_{bb}) \}$
            \Comment{Perform best grasp.}
        \Else
            \State $a \gets \{ \textbf{view: } \argmax(\mathcal{M}) \}$ \Comment{Select informative view}
            \State $A_{prev} \gets A$ 
            \State $\mathcal{E} \gets \text{PerformAction}(\mathcal{E}, a$
            \Comment{Update view}
            \State $A, \mathcal{C}_{bb} \gets \text{AdjNet}(\mathcal{E})$
            \Comment{Run Adj-Net on new view}
            \State $A \gets \text{FuseAdjacency}(A_{prev}, A)$ \Comment{Posterior adj. mat.}
        \EndIf
    \EndWhile
    \newline
    \Return $a$ \Comment{Returns the best grasping action.}
\end{algorithmic}
\end{algorithm}

	\section{Experiments}
	\label{sec:experiments}
	
\subsection{Implementation Details}

In our implementation, Adj-Net utilizes Deformable DETR \cite{Zhu20arxiv} with  ResNet101 as the feature extractor. Compared to using DETR and/or ResNet50, this combination yields the best results in all metrics (see Sec.~\ref{sec:AdjNet_SOTA}). For the action-selection experiment, Adj-Net was trained on the VMRD dataset \cite{Zhang18ichr} modified to only consider binary classification, i.e., candidate classes "object" and "unknown". This modification yields better performance in object relationship metrics and faster training than when using all VMRD's 31 classes + "unknown" class. Furthermore, explicit knowledge of the object class is not required in our real-world experiments. For a benchmark comparison with object stacking state-of-the-art algorithms, Adj-Net was also trained on the full 31+1 object classes of VMRD. We initiated our training from a pre-trained Deformable DETR on COCO17 \cite{Lin14eccv}, with $q=300$ bounding box queries and a hidden dimension of $h=256$. Adj-Net was trained for 1500 epochs with a batch size of 8, using a system with two Nvidia RTX A5000 GPUs. The training was done in two stages. First, we trained the detector only by setting $\gamma_{adj} = 0$ using Stochastic Gradient Descent optimizer and Super-convergence scheduler with a maximum learning rate of 0.01 for 800 epochs. Next, we activated the learning of the adjacency matrix and used the AdamW optimizer with a maximum learning rate of 0.001. In both phases, we used a weight decay of $10^{-6}$. In our configuration, Adj-Net's architecture contains around 41M and 64M trainable parameters for ResNet50 and ResNet101 versions, respectively.

For depth estimation, we used a pre-trained DPT-Net \cite{Ranftl21iccv} with the DPT-Large dataset. To generate grasp quality maps, we used a pre-trained GR-ConvNet \cite{Kumra2020iros}, trained on the Jacquard dataset \cite{Depierre18iros}. To decide, if needed, the next viewpoint from which to take an image, we utilized the MVP Controller \cite{Morrison19icra}. The data-association between objects from different images was performed using Superpoint \cite{Detone18cvpr} for keypoint detection and using Superglue \cite{Sarlin20cvpr} to match the keypoints. The matching between bounding boxes was selected by choosing the boxes with the highest number of matching keypoints. 

\subsection{Adj-Net Performance}\label{sec:AdjNet_SOTA}

We compared the performance of our algorithm on the VMRD dataset against current state-of-the-art approaches for object-stacking detection: Multi-task CNN \cite{Zhang19iros}, VMRN \cite{Zhang20prl} and GVMRN \cite{Zuo21ijcnn}. We considered a relationship inference to be correct if:

\begin{itemize}
    \item For objects $i$ and $j$ where $i$ is placed over $j$, $\mathbb{P}(\exists \epsilon_{i \rightarrow j}|I) > 0.5$ and $\mathbb{P}(\exists \epsilon_{i \rightarrow j}|I) > \mathbb{P}(\exists \epsilon_{j \rightarrow i}|I)$.
    \item For objects $i$ and $j$ without a direct relation between them, $\mathbb{P}(\exists \epsilon_{i \rightarrow j}|I) < 0.5$ and $\mathbb{P}(\exists \epsilon_{j \rightarrow i}|I) < 0.5$.
\end{itemize}
For comparison, we also considered Adj-Net trained for binary classification.
We considered three standard benchmarks for stacking relationship inference:
\begin{itemize}
    \item \emph{Object Precision (OP)}: The number of true positive detected relationships divided by the total number of detected relationships. A detected relationship is considered a true positive if the tuple $(C_i, R_{ij}, C_j)$ is correct, where $C_\square$ denotes the $\square$th object class, and $R$ the relationship between the two objects in the subscript.
    \item \emph{Object Recall (OR)}: The number of true positive detected relationship divided by the number of ground truth relationships. The definition of a true positive detection is the same as for OP. 
    \item \emph{Image Accuracy (IA)}: The percent of images in the test set for which OP and OR are $100\%$, i.e., all the objects detected with their correct classes and relationships without false positives. We display additional results for image accuracy with a specific number of objects as IA-x, where x represents the number of objects in the scene.
\end{itemize}

The results are displayed in Tables 
\ref{tab:Adj-Net_Performance} and \ref{tab:Adj-Net_IA}. Even with our smaller Adj-Net variant with the ResNet50 feature extractor, our approach outperforms the current state-of-the-art \VT{in OR, OP, and AI metrics}, listed in table~\ref{tab:Adj-Net_Performance}. When examining IA for a specific number of objects per image in Table~\ref{tab:Adj-Net_IA}, the weaker feature extractor plays a factor in the relatively low accuracy for images with 2 objects. In contrast, IA for images with a larger number of objects shows superior accuracy and consistency, as IA does not drop as the number of objects increases. The larger variant with ResNet101 generates the best current results for VMRD because of the stronger backbone. Binary classification versions suffer less from classification errors and, thus, are able to reach very high scores all-around. Thus, for applications that do not require accurate object classification, such as bin picking, binary classification versions are preferred. 


\begin{table}[]
\small
    \begin{center}
        \begin{tabular}{ c|ccc }
         \hline
         \textbf{Models}  & \textbf{OR} & \textbf{OP} & \textbf{IA} \\ 
         \hline
         Multi-task CNN \cite{Zhang19iros}       & 86.0  & 88.8  & 67.1 \\ 
         VMRN-RN101 \cite{Zhang20prl}           & 85.4  & 85.5  & 65.8  \\
         VMRN-VGG16 \cite{Zhang20prl}           & 86.3  & 88.8  & 68.4  \\
         GVMRN-RF-RN101 \cite{Zuo21ijcnn}       & 86.9  & 87.5  & 68.8 \\
         GVMRN-RF-VGG16 \cite{Zuo21ijcnn}       & 88.7  & 89.5  & 70.2 \\
         \hline
         \VT{\textbf{Adj-Net RN50 (Ours)}}                  & 88.9  & \textbf{91.5} & 75.0  \\
         \VT{\textbf{Adj-Net RN101 (Ours)}}                 & \textbf{89.8}  & \textbf{91.5}  & \textbf{77.3}  \\
         \hline
         \hline
         \textbf{Adj-Net Binary RN50 (Ours)}           & 90.1  & 91.6  & 77.5  \\
         \textbf{Adj-Net Binary RN101 (Ours)}          & 89.8  & 92.3  & 79.0  \\
         \hline
        \end{tabular}
    \end{center}
    \caption{Benchmark comparison on the VMRD dataset of state-of-the-art approaches with ours in the OR, OP and IA metrics. The binary Adj-Net variations are separated as they consider binary classification instead of the full (31 classes) classification.}
    \label{tab:Adj-Net_Performance}
\end{table}

\begin{table}[]
\small
    \begin{center}
        \begin{tabular}{ c|cccc }
         \hline
         \textbf{Models} & IA-2 & IA-3 & IA-4 & IA-5\\ 
         \hline
         Multi-task CNN \cite{Zhang19iros}       & 87.7 & 64.1 & 56.6 & 72.9 \\ 
         GVMRN-RF-RN101 \cite{Zuo21ijcnn}        & 91.4 & 69.2 & 61.2 & 57.5\\
         GVMRN-RF-VGG16 \cite{Zuo21ijcnn}        & \textbf{92.9} & 70.3 & 63.8 & 60.3\\
         \hline
         \VT{\textbf{Adj-Net RN50 (Ours)}}                   & 87.3 & 74.5 & 69.8 & 72.6   \\
         \VT{\textbf{Adj-Net RN101 (Ours)}}               & 88.7 & \textbf{75.2} & \textbf{75.0} & \textbf{76.7} \\
         \hline
         \hline
         \textbf{Adj-Net Binary RN50 (Ours)}            & 92.9 & 75.6 & 71.5 & 78.0 \\
         \textbf{Adj-Net Binary RN101 (Ours)}           & 88.7 & 79.4 & 73.2 & 78.0\\
         \hline
        \end{tabular}
    \end{center}
    \caption{Benchmark comparison on the VMRD dataset of state-of-the-art approaches with ours in the IA-2 to IA-5 metrics. The VMRN algorithms' benchmarks are not presented by Zhang et al. \cite{Zhang20prl}. The binary Adj-Net variations are separated as they consider binary classification instead of the full (31 classes) classification.}
    \label{tab:Adj-Net_IA}
\end{table}


\subsection{Robot Experiment Setup}

\begin{figure}
    \begin{center}
        \includegraphics [width=0.9\textwidth]{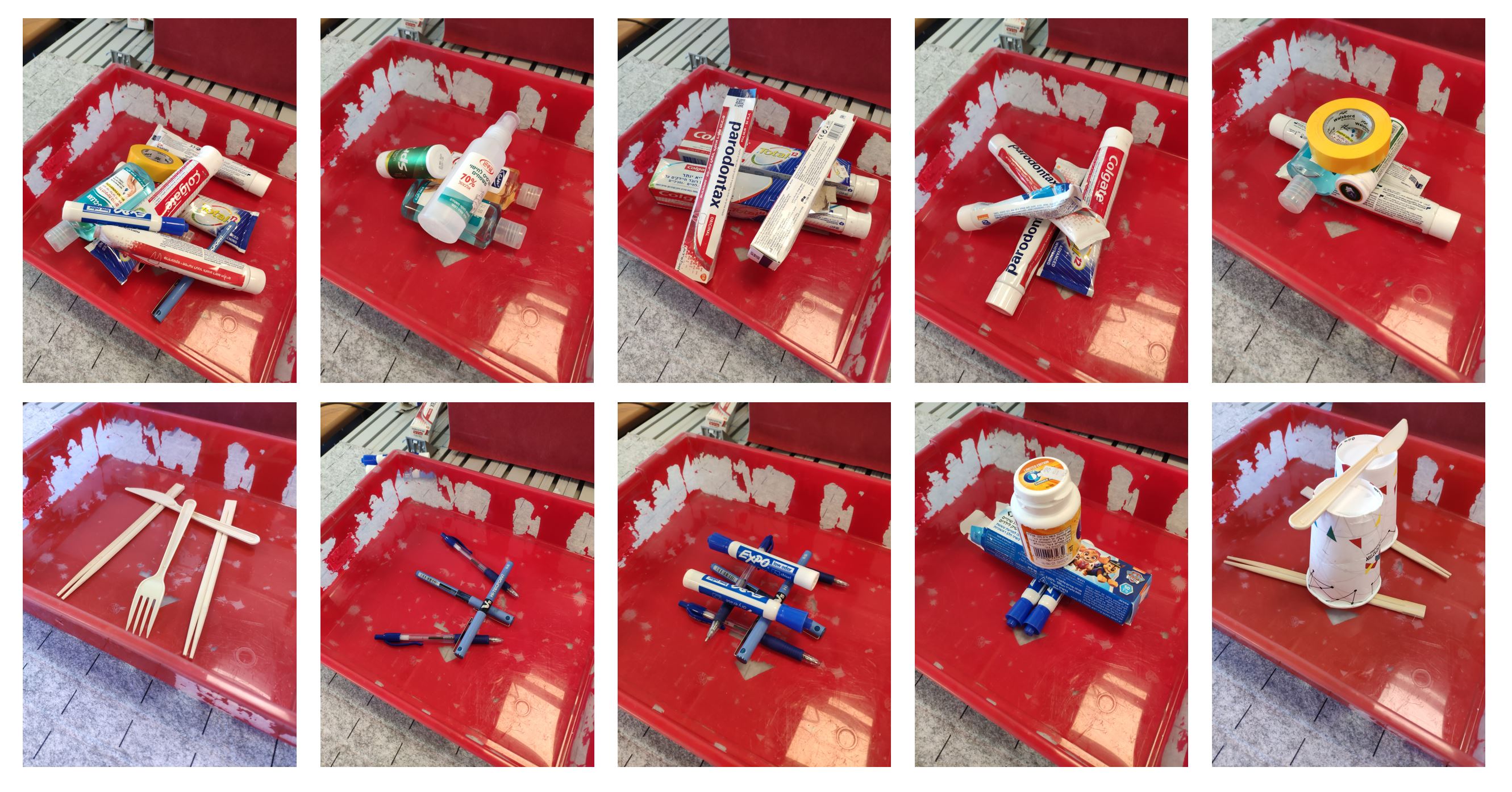}
    \end{center}
    \vspace{-0.5cm}
    \caption{The ordered scenarios considered in our robot experiments.}
    \label{fig:Scenarios_exp}
\end{figure}

\begin{figure}
    \begin{center}
        \includegraphics [width=0.9\textwidth]{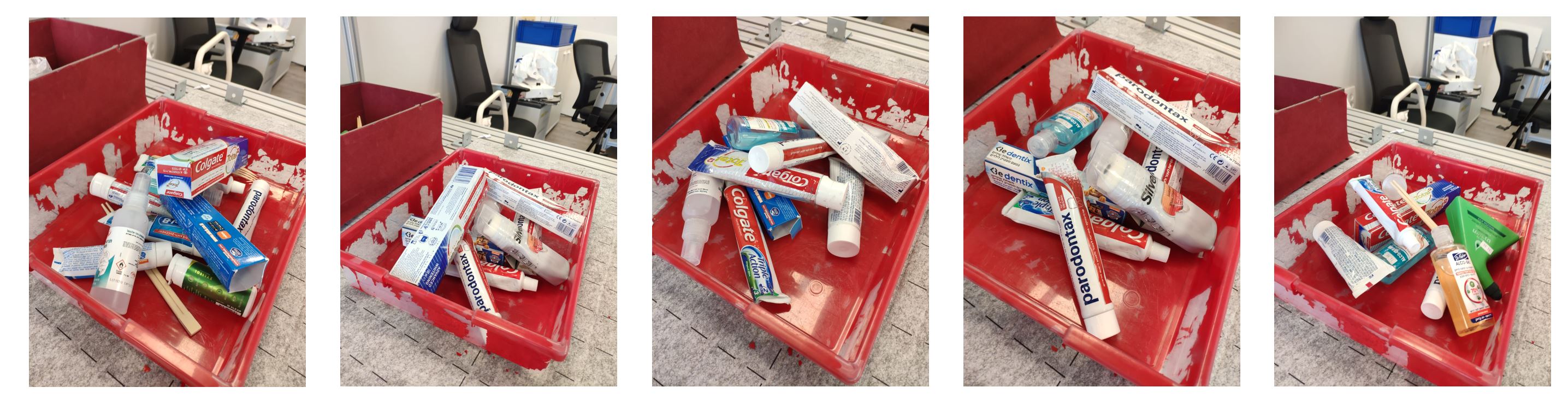}
    \end{center}
    \vspace{-0.5cm}
    \caption{The random scenarios considered in our robot experiments.}
    \label{fig:Scenarios_rand}
\end{figure}

We use the Franka Emika Panda for our grasping experiment. We implement the action-selection scheme presented in Sec.~\ref{sec:Action_Select}. The maximum information entropy parameter considered for the experiment is $\mathcal{H}_{th} = 0.45$. We perform bin picking in a setting with various types of objects in various setups: office supplies, kitchen tools, various paste tubes and liquid bottles, and package boxes, all presented in Fig.~\ref{fig:Scenarios_exp}. These scenarios, \VT{denoted 'Orderly stack',} represent settings of stacked objects where accurate grasping is required to avoid \emph{spillage} or \emph{breaking}, and are designed to challenge approaches that do not reason about object hierarchy. For this experiment, we set the robot's gripper to apply little force so not to break the objects or spill the liquid contained in them. We compare DUQIM-Net to a baseline of a grasping approach that only uses the grasping quality map values created by GR-ConvNet.
\VT{We also present average results for additional scenarios with more objects where they are randomly stacked in a pile, denoted 'Random stack', showed in Fig.~\ref{fig:Scenarios_rand}}.

All models in use  are pre-trained. Specifically AdjNet was trained on the VMRD dataset and did not see the objects' configuration in the scenario, showcasing its ability to generalize to previously unseen scenarios. This saves precious learning time in the lab, as the training was done offline.
We ran each approach 5 times per scenario for 10 scenarios while observing the operation, amounting to 50 runs for each approach. In the following section, we present the average results of our benchmarks.


\subsection{Robot Experiment Results}

We present in Table \ref{tab:Exp_Bin_Clearing} the results for the bin picking. Shown are the average number of grasp attempts, grasp successes and additional views taken. In addition, we consider the following benchmarks:
\begin{itemize}
    \item \emph{Grasp Success:} object-grasping success rate.
    \item \emph{Action Efficiency:} the ratio between the number of successful grasps to the total number of actions: grasping attempts, pushes, and taking additional viewpoints.
    \item \emph{Object Order Error (OOE):} the ratio between interactions with objects that support other objects (incorrect grasping order) and the number of pushing \& grasping attempts.
\end{itemize}

Our overall grasping capability for individual objects is determined by GR-ConvNet. Thus, we run the experiment on scenarios that can be run to completion (i.e., excluding adversarial objects for grasping), rendering the common \emph{completion} and \emph{clearance} metrics for grasping irrelevant.

When training Adj-Net Binary only on the VMRD dataset, our network is able to generalize to unseen scenarios and objects in the dataset and improve upon the baseline where Adj-Net is not used. We show significant improvements in OOE, with almost three times \VT{and two times} less grasps of an object in an incorrect order \VT{for the ordered and random scenarios respectively}. In addition, we indirectly improve the grasping success by preventing the grasped objects from being pulled down by the objects they were supporting, which is significant in a setting where the grasp does not apply a strong force, as in our experiment. This improvement comes at the cost of action efficiency, as sometimes DUQIM-Net requires additional viewpoints to determine the correct object hierarchy.

\begin{table*}
    \begin{center}
    \small{
        \begin{tabular}{c|c|ccc|ccc} 
         \hline
         \textbf{Scenario} & \textbf{Method} & \textbf{$\downarrow$ Grasp Att.} & \textbf{Grasp Suc.} & \textbf{Views Added} & \textbf{$\uparrow$Grasp Suc.\%} & \textbf{$\uparrow$Action Eff.\%} & \textbf{$\downarrow$OOE.\%}\\ 
         \hline
         Orderly stack & GR-ConvNet     & 6.14 & 5.0 & 0 & 85.2 & \textbf{85.2} & 24.6 \\
         & Ours           & \textbf{5.76} & 5.10 & 2.08 & \textbf{91.1} & 71.5 & \textbf{10.0} \\
         \hline
         \VT{Random Stack }& GR-ConvNet     & 13.4 & 10.2 & 0 & 77.5 & \textbf{77.5} & 27.0 \\
         & Ours           & \textbf{11.6} & 10.2 & 7.80 & \textbf{88.6} & 53.5 & \textbf{15.3} \\
         \hline
        \end{tabular}}
    \end{center}
    \caption{Average results for the bin clearing task on \VT{both our scenario types}. \emph{OOE} (object order error) represents the percent of scenarios where a supporting object was grasped. The difference in the number of successful grasps is due to the occasional grasp of two objects at once. The arrow direction represents which direction is better, up or down for higher or lower values, respectively.}
    \label{tab:Exp_Bin_Clearing}
\end{table*}


	\section{Conclusions}
	\label{sec:conclusions}
	We presented DUQIM-Net, a system for object manipulation in cluttered settings. When deciding between grasping an object or taking an additional image from a different viewpoint, DUQIM-Net considers the hierarchical underlying structure of the object clutter. At its core is the Transformer Encoder-Decoder-based Adj-Net, which infers from a given an RGB image the object hierarchy in the form of a weighted adjacency matrix. This flexible representation is used in our proposed action-selection algorithm and is subsequently applied to a real robot setting. In our experiments, we show that Adj-Net outperforms current CNN-based state-of-the-art approaches when it comes to inferring object hierarchy on the VMRD dataset. In addition, we demonstrate, via bin picking, the advantage of using object hierarchy information and the generalization ability of the binary classification Adj-Net.
	
	\section{Acknowledgements}
	\label{sec:acknowledgements}
	The authors would like to thank Zohar Feldman and Oren Spector from BCAI for their support and numerous useful discussions.

	
	

	\bibliographystyle{IEEEtran}
	\bibliography{refs}

\end{document}